\RequirePackage{fix-cm}

\documentclass[smallextended]{svjour3}  

\smartqed 
\usepackage{graphicx}

\usepackage{amssymb}
\usepackage{latexsym}

\usepackage{url}
\usepackage{xcolor}
\definecolor{newcolor}{rgb}{.8,.349,.1}

\usepackage{times}
\usepackage{epsfig}
\usepackage{balance}
\usepackage{subfigure}
\usepackage{enumitem}
\setlist{nosep}
\usepackage{booktabs}
\usepackage{tabularx}
\usepackage{multirow}

\usepackage{tabularx}
\usepackage{cite}
\usepackage{balance}
\usepackage{subfigure}
\usepackage[multiple]{footmisc}
\usepackage{float}
\usepackage{stfloats}
\usepackage{multirow}
\usepackage{hyperref}
\usepackage{amssymb}
\usepackage{pifont}

\usepackage{cite}

\usepackage{xspace}
\makeatletter
\DeclareRobustCommand\onedot{\futurelet\@let@token\@onedot}
\def\@onedot{\ifx\@let@token.\else.\null\fi\xspace}
\def\eg{\emph{e.g}\onedot} 
\def\ie{\emph{i.e}\onedot}

\def \etal{\emph{et~al}\onedot}
\makeatother

\usepackage{pifont}
\newcommand{\cmark}{\ding{51}}%

\newcommand{\highlight}[1]{\textcolor{black}{#1}}

\begin{document}
%
% \title{\highlight{Context-based Instance Segmentation for the DAVIS Challenge on Semi-Supervised Video Object Segmentation}}

\title{\highlight{Contextual Guided Segmentation Framework for Semi-supervised Video Instance Segmentation}}

% \author[1]{Trung-Nghia Le}
% \author[2]{Tam V. Nguyen}
% \author[3]{Minh-Triet Tran}

% \affil[1]{National Institute of Informatics, Tokyo, Japan}
% \affil[2]{Department of Computer Science, University of Dayton, Ohio, United States of America}
% \affil[3]{University of Science, VNU-HCM, Ho Chi Minh, Vietnam}

\author{Trung-Nghia Le         \and
        Tam V. Nguyen         \and
        Minh-Triet Tran       
}

\institute{Trung-Nghia Le \at
              National Institute of Informatics, Tokyo, Japan.
              \email{{\tt\small ltnghia@nii.ac.jp}}           %  \\
           \and
           Tam V. Nguyen \at
              Department of Computer Science, University of Dayton, Ohio, USA.
             \email{{\tt\small tamnguyen@udayton.edu}}   
            \and
           Minh-Triet Tran \at
              Corresponding author, University of Science and Vietnam National University, Ho Chi Minh, Vietnam.
              \email{{\tt\small tmtriet@fit.hcmus.edu.vn}}   
}

\date{Received: date / Accepted: date}
\maketitle              % typeset the header of the contribution
\begin{abstract}
\highlight{In this paper, we propose Contextual Guided Segmentation (CGS) framework for video instance segmentation in three passes.} In the first pass, \ie  preview segmentation, we propose Instance Re-Identification Flow to estimate main properties of each instance (i.e., human/non-human, rigid/deformable, known/unknown category) by propagating its preview mask to other frames. \highlight{In the second pass, \ie contextual segmentation, we introduce multiple contextual segmentation schemes. For human instance, we develop skeleton-guided segmentation in a frame along with object flow to correct and refine the result across frames. For non-human instance, if the instance has a wide variation in appearance and belongs to known categories (which can be inferred from the initial mask), we adopt instance segmentation. If the non-human instance is nearly rigid, we train FCNs on synthesized images from the first frame of a video sequence. In the final pass, \ie guided segmentation, we develop a novel fined-grained segmentation method on non-rectangular regions of interest (ROIs). The natural-shaped ROI is generated by applying guided attention from the neighbor frames of the current one to reduce the ambiguity in the segmentation of different overlapping instances. Forward mask propagation is followed by backward mask propagation to further restore missing instance fragments due to re-appeared instances, fast motion, occlusion, or heavy deformation. Finally, instances in each frame are merged based on their depth values, together with human and non-human object interaction and rare instance priority. Experiments conducted on the DAVIS Test-Challenge dataset demonstrate the effectiveness of our proposed framework. We achieved the \textbf{$3^{rd}$} consistently in the DAVIS Challenges 2017-2019 with 75.4\%, 72.4\%, and 78.4\% in terms of global score, region similarity, and contour accuracy, respectively.}

\end{abstract}
%
%
%
%%%%%%%%% BODY TEXT
\section{Introduction}

\begin{figure}[!t]
    \begin{center}
  \includegraphics[width=\linewidth]{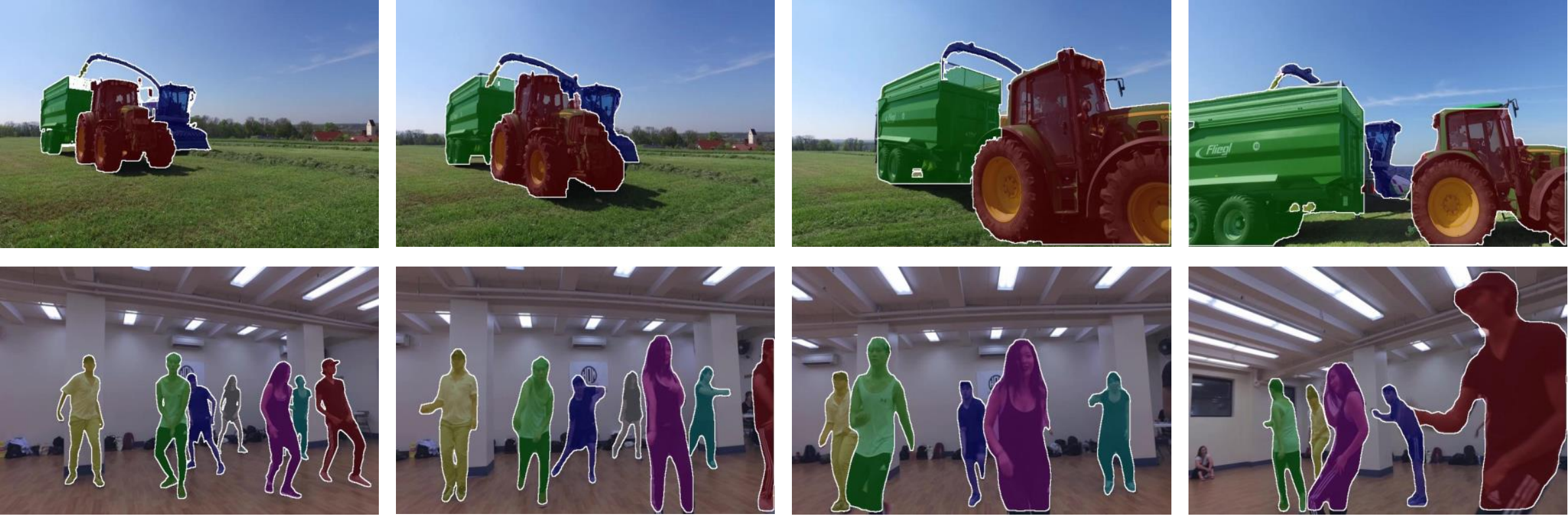}
        \end{center}
        \centering
        %\vspace{-6mm}
    \caption{Examples of results obtained by our proposed method. From left to right: the first video frame with the ground-truth label followed by results of our method on next frames.}
    \label{fig:examples}
\end{figure}

Object segmentation is considered a labeling problem aiming to separate foreground from background regions. Video instance segmentation, which is higher-level and more challenging than object segmentation, aims to label each video frame pixel to instances or the background region and then assign consistent IDs to these instances over the video sequence. Object/instance segmentation in videos is beneficial in a wide range of practical applications, \ie, autonomous vehicle~\cite{Brabandere-CVPRW2017}, action recognition~\cite{Ji-ECCV2018}, video summarization~\cite{Lee-IJCV2015}, object tracking~\cite{Wang-CVPR2019}, scene understanding~\cite{Xiong-CVPR2019}, and video annotation~\cite{ltnghia-AAAI2021}.

This paper focuses on semi-supervised video instance segmentation~\cite{Jordi-2017}, which targets certain instances whose ground-truth mask for the first video frame is given. DAVIS Challenge~\cite{Jordi-2017} promotes the development of this task. The benchmark dataset of this challenge consists of many pitfalls such as rapid motion, distractors, smaller objects, fine structures, occlusions, large deformations, complex object interactions, and so on.  Figure \ref{fig:examples} shows some exemplary results of our proposed method on the DAVIS Test-Challenge dataset~\cite{Jordi-2017}.

\highlight{To address the challenges of the given problem, tracking and re-identification methods are adopted and jointly integrated into segmentation models to keep the consistency of targeted instances over the entire video sequence~\cite{Khoreva-DAVIS2017, Li-DAVIS2017, Guo-DAVIS2018, Li-DAVIS2018}. However, existing works usually fail to follow and segment targeted instances due to cannot cover all various contexts in the video. We argue that context information is essential for semantic segmentation to reduce ambiguous instances and obtain robust results. Therefore, this work aims to leverage the context information to improve the performance of video instance segmentation. Inspired by the idea of \textit{"you should look twice"}~\cite{YADA,YALA} in the task of object detection, we propose a three-pass guided segmentation framework, namely Contextual Guided Segmentation (CGS), to tackle the problem of semi-supervised video instance segmentation. Our proposed method consists of two key ideas as below.}

\highlight{First, we exploit variation in the video and propose various contextual segmentation strategies adapting to contexts, \ie the category and visual properties of an instance. To select the appropriate scheme, we propose a novel Instance Re-Identification Flow (IRIF) to propagate the initial mask of an instance to other frames and analyze the visual properties of segmented regions. Multiple contextual segmentation schemes are also introduced to adapt the contextual properties of each instance. For human instances, we develop skeleton-guided segmentation. For non-human instances, we train FCNs from our synthesized dataset for nearly-rigid instances with similar background scenes. Instance segmentation detectors are utilized to handle deformable non-human instances in known categories. Results from our IRIF are treated as the baseline scheme for other cases.}

\highlight{Second, to segment an instance in a region of interest (ROI), we propose novel guided fined-grained segmentation based on attention for performance improvement. We transform a regular rectangular ROI to a non-rectangular ROI by blending attention inferred from neighbor frames to eliminate complex background inside the ROI. We also propose bi-directional propagation strategies to construct adaptive attention for guided segmentation. Forward propagation strategy can correct missing segmentation due to dense objects in a ROI. Meanwhile, a backward propagation strategy can recover missing instances due to fast motion, occlusion, or heavy deformation. }

\highlight{The DAVIS Challenges 2017-2019 results indicate that our method is competitive among the top-performing submissions. Our early results were preliminarily listed on DAVIS 2017 Challenge~\cite{Le-DAVIS2017}, DAVIS 2018 Challenge~\cite{tmtriet-DAVIS2018}, and DAVIS 2019 Challenge~\cite{tmtriet-DAVIS2019}. In this paper, we provide the full details of our proposed framework. Our contributions are as follows.}

\begin{itemize}
    \item \highlight{We propose Contextual Guided Segmentation (CGS) framework with three segmentation passes to exploit various contexts in video instance segmentation. Our proposed method achieved the \textbf{$3rd^{th}$} ranking consistently in the DAVIS Challenges 2017-2019.}
    
    \item We propose Instance Re-Identification Flow (IRIF) to extract contextual properties of each instance by propagating its preview mask from the current frame to coming frames.
    \item We introduce multiple contextual segmentation schemes to adapt the contextual properties of each instance.
    
    \item \highlight{We propose bi-directional propagation strategies for guided fined-grained segmentation in non-rectangular ROIs. Our proposed guided segmentation outperforms the standard segmentation, which is mostly applied in rectangular ROIs.}
    \item To blend instance masks into a unique result, we introduce a merging process based on their depth values together with human and non-human object interaction and rare instance priority.
    
    \item \highlight{We construct Wonderland Data to increase the number of training data for one-shot learning. Our proposed augmentation approach also can be utilized for different problems.}

\end{itemize}

The remainder of this paper is organized as follows. In Section~\ref{sec:related_work}, we briefly review the related work. Next, our proposed methods are presented in Section~\ref{sec:ProposedMethod}. Experimental results are then reported and discussed in Section~\ref{sec:results}. Finally, Section~\ref{sec:conclusion} concludes and paves the way for future work.

%-------------------------------------------------------------------------

\section{Related Work}
\label{sec:related_work}

\subsection{One-Shot Learning}

Data augmentation is essential to deal with one-shot learning~\cite{Caelles-CVPR2017}, which aims to train a deep network with only a given first video frame. Caelles \etal ~\cite{Caelles-CVPR2017} introduced the first simple data augmentation strategy such as random crop, random scale, vertical flip, random changes in brightness, saturation, and contrast of the given first frame. Khoreva \etal~\cite{Khoreva-DAVIS2017} later introduce Lucid Dreaming~\cite{Khoreva-DAVIS2017} to synthesize the foreground changes by rigid and non-rigid transformation with a small extent, and synthesize the background changes using affine deformations with limited appearance variations. The given first frame with ground truth is augmented with Lucid Dreaming to generate more training data with different viewpoints, leading to much improvement of training networks. Hence, augmented data by Lucid Dreaming, called Lucid Data, has become common for one-shot learning. However, Lucid Data cannot deal with different backgrounds caused by objects' motion or camera view changes. \highlight{Guo \etal~\cite{Guo-DAVIS2018} changed the background of the first video frame by images with pure background crawled randomly from the Internet by Google, namely Online Data. However, Online Data is unstable because of randomly crawled from the Internet without considering the content of the video. Meanwhile, our Wonderland Data is filtered out from large-scale scene data to choose the most similar scenes with the video. }

Khoreva \etal~\cite{Khoreva-DAVIS2017} trained appearance-based and motion-based models with Lucid Data~\cite{Khoreva-DAVIS2017}. Shaban \etal~\cite{Shaban-DAVIS2017} learned video segments by bootstrapping them from temporally consistent object proposals, which are first spatially trained on Lucid Data~\cite{Khoreva-DAVIS2017} and then incorporated a semi-Markov pixel-level motion model to form spatio-temporal object proposals. Luiten \etal~\cite{Luiten-DAVIS2018} first trained DeepLab3+~\cite{Chen-ECCV2018} on a combination of standard datasets and then fine-tuned the network on Lucid Data~\cite{Khoreva-DAVIS2017} of each video to form a strong network to segment instance inside ROI. Li \etal~\cite{Li-DAVIS2018} trained online re-identification network, which is the original Region Proposal Network of Mask R-CNN, and a recurrent mask propagation network on Lucid Data~\cite{Khoreva-DAVIS2017}. Xu ~\cite{Kai-CVPR2019} proposed a spatio-temporal CNN in which the spatial segmentation branch is fine-tuned online on Lucid Data of each sequence while the temporal coherence branch is trained offline on the entire dataset. Models are not only fine-tuned offline on Lucid Data~\cite{Khoreva-DAVIS2017} of the first frame but also can be updated online while processing the video~\cite{Voigtlaender-DAVIS2017}. Mask R-CNN is fine-tuned on Lucid Data~\cite{Xu-CVPR2019} or Online Data~\cite{Guo-DAVIS2018} to adapt proposals to the video.

%%%%%%%%%%%%%%%%%%%

\subsection{Temporal Connection Mining}

This approach aims to perform instance tracking, propagation, and re-identification, where each instance is detected and re-identified through frames~\cite{Li-DAVIS2017}. Li \etal~\cite{Li-DAVIS2017} iteratively propagated masks via flow warping and re-identified instances via adaptive matching to retrieve missing ones. Luiten \etal~\cite{Luiten-DAVIS2018} first segmented multiple object proposals in the entire video and then selected and linked these proposals over time using a re-identification feature embedding vector for each proposal. Re-identification feature embedding vectors are computed using a triplet-loss based re-identification embedding network. Li \etal~\cite{Li-DAVIS2018} jointed re-identification and attention-based recurrent temporal propagation into a unified framework to retrieve missing objects despite their large appearance changes. Guo \etal~\cite{Guo-DAVIS2018} first extracted possible mask proposals in each frame and then joined tracking and re-identification to filter and rank proposals to merge the highest confident proposals. Xu \etal~\cite{Xu-CVPR2019} adapted a multiple hypotheses tracking method to build up a bounding box proposal tracking tree for different objects, then propagate masks, and finally merged mask proposals from the tracking tree. Wang \etal~\cite{Wang-CVPR2019} used fully convolutional Siamese trackers to produce class-agnostic binary segmentation masks of the target objects. Voigtlaender \etal~\cite{Voigtlaender-CVPR2019} used a semantic pixel-wise embedding together with a global and a local matching mechanism to transfer information from the first frame and from the previous frame of the video to the current frame, which is used as internal guidance for segmentation. \highlight{Jonathon \etal~\cite{Luiten-DAVIS2019} used a Siamese architecture to detect and track multiple objects and then performed segmentation inside the detected bounding boxes. Tran \etal~\cite{tmtriet-DAVIS2020} propagated masks with reference to multiple extra samples through a memory reference pool.}

\subsection{End-to-End Temporal Learning}

\highlight{This approach directly learns temporal information in a video through deep learning architectures such as LSTM, guided-attention, or memory networks. Some methods combine feature maps from different video frames by correlation matching~\cite{Voigtlaender-CVPR2019} or non-local matching~\cite{Oh-DAVIS2019}. Guo \etal~~\cite{Guo-DAVIS2019} integrated STM~\cite{Oh-DAVIS2019} into DeepLabv3+~\cite{Chen-ECCV2018} to concatenate low-level features in mask decoder. Andreas \etal~\cite{Robinson-CVPR2020} implemented a memory network to add semantic information about the target object from a previous frame to the refinement stage, complementing the predictions provided by the target appearance model. Zhang \etal~\cite{Zhang-DAVIS2020} developed a spatial constraint module that takes the previous prediction to generate a spatial prior for the current frame, helping to disambiguate appearance confusion and eliminate false predictions. Fiaz \etal~\cite{Fiaz-DAVIS2020} introduced a guided feature learning without model update algorithm for directional deep appearance learning. Liu \etal~\cite{Liu-DAVIS2020} integrated multilevel backbone into memory network to generate higher spatial resolution features. Le \etal~\cite{vltanh-DAVIS2020} leveraged existing memory-based models and enhanced their capability by adding pre-processing and post-processing steps. Xie \etal~\cite{Xie-DAVIS2020} integrated depth maps from a video
sequence into STM~\cite{Oh-DAVIS2019} to alleviate the ambiguity of objects with similar appearances. Seong \etal~\cite{Seong-DAVIS2020} developed a kernelized memory network and used the Hide-and-Seek strategy training to handle occlusions and segment boundary extraction. Yang \etal\cite{Yang-DAVIS2020} combined collaborative foreground-background integration with multi-scale matching to be robust to various object scales.}

%-------------------------------------------------------------------------

\section{Proposed Method}
\label{sec:ProposedMethod}

\subsection{\textbf{Overview}}

\begin{figure}[t]
    \centering
    \includegraphics[width=1\textwidth]{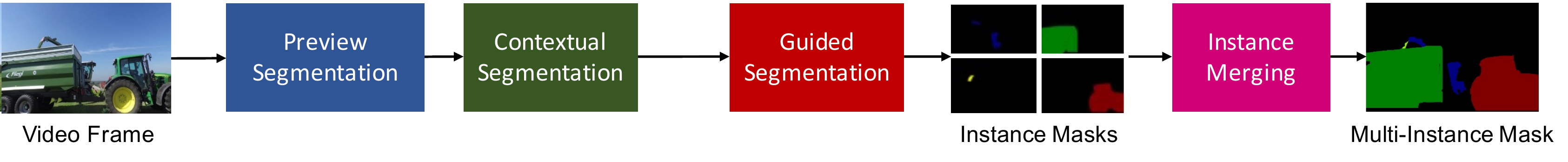}
    \caption{\highlight{Overview of our Contextual Guided Segmentation (CGS) framework.}}
    \label{fig:cgs}
\end{figure}

\highlight{Figure \ref{fig:cgs} illustrates CGS with three passes: preview segmentation for context evaluation, contextual segmentation, guided segmentation based on propagation. In particular, In the first pass, we propose Instance Re-Identification Flow (IRIF) to generate the preview mask sequence and extract different contextual properties from each instance. In the second pass, we introduce multiple segmentation schemes corresponding to extracted properties. In the third pass, we develop fined-grained segmentation based on guided propagation. We remark that each instance is processed independently over frames of a video sequence. Finally, instance masks are then blended with reference to depth information, human and non-human instance interaction, and rare instance priority.}

\subsection{\textbf{Preview Segmentation}}

\begin{figure}[t]
    \begin{center}
  \includegraphics[width=0.8\textwidth]{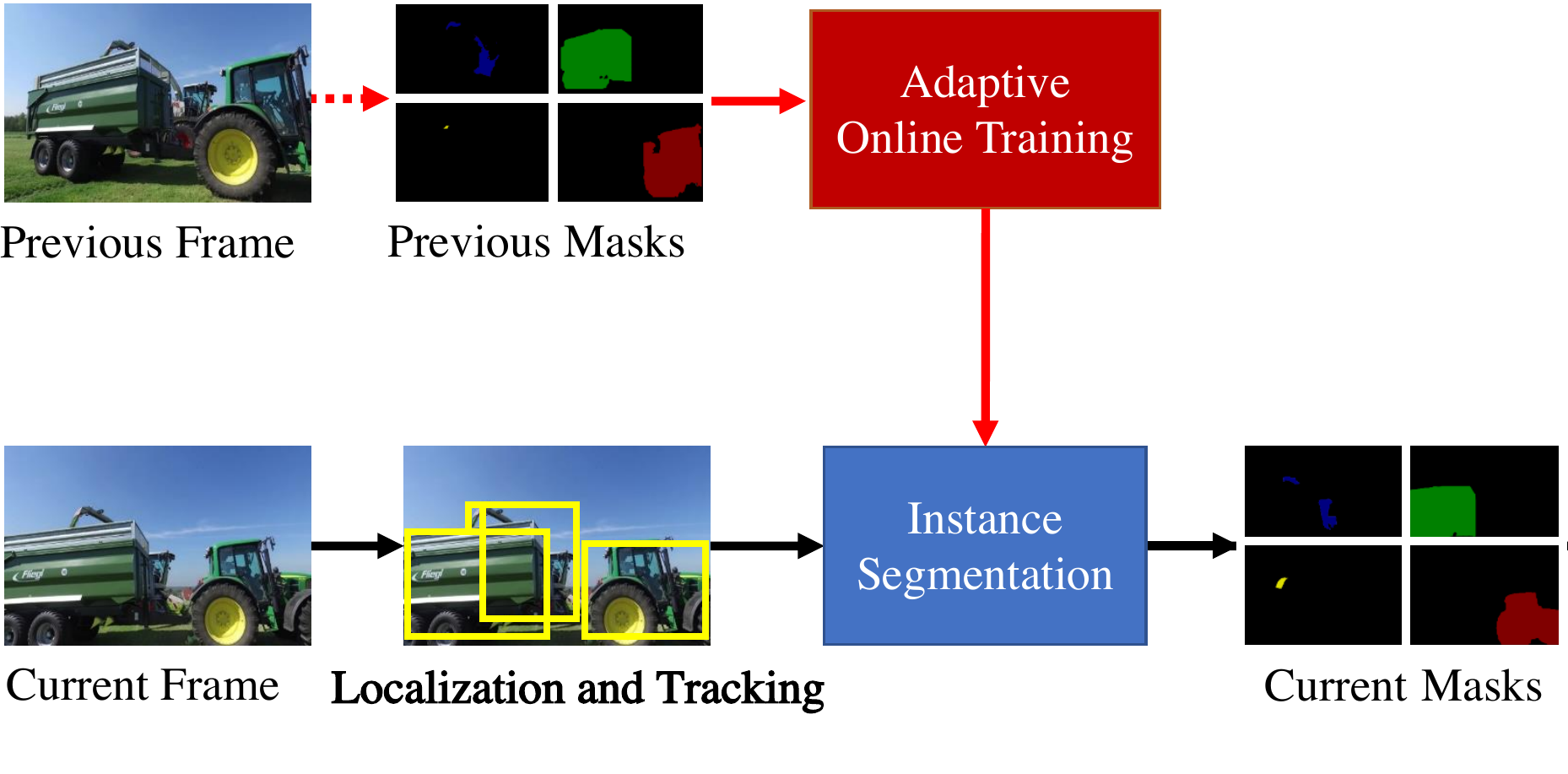}
        \end{center}
        \centering
    \caption{\highlight{The flowchart of Instance Re-Identification Flow (IRIF) component. The segmentation performed on the current frame is based on the history information of the previous frames. The segmentation result of the current frame is further fed to the process of the coming frame.}}
    \label{fig:IRIF}
\end{figure}

Figure~\ref{fig:IRIF} illustrates the flow chart of Instance Re-Identification Flow (IRIF) for preview segmentation. The segmentation performed on the current frame is based on the history information of the previous frames. The segmentation result of the current frame is further fed to the process of the coming frame.

We remark that in this component, we consider two types of instance, \ie~ human and non-human, to treat each instance in different ways. Given the first frame with its ground truth label, we extract the bounding box for each instance and then perform human/non-human classification for all instances using Mask R-CNN~\cite{Kaiming-ICCV2017}.

\subsubsection{Instance Localization and Tracking}

For each video frame, we localize and track instances in a re-identification manner. Note that we expand the bounding box to 10\% to well capture the whole area of the object instances. \highlight{For \textit{human objects}, we employ person search~\cite{Tong-CVPR2017} by detecting person by using Faster R-CNN and then extracting person re-identification feature for all detected person region.} On the other hand, DeepFlow~\cite{Weinzaepfel-ICCV2013} and Deformable Part Models (DPM)~\cite{Pedro-CVPR2008} are utilized to detect and track \textit{non-human objects}.  

\subsubsection{Adaptive Online Learning for Instance Segmentation}

For each instance, to identify each pixel as foreground (instance) or background, we utilize multiple binary SVM classifiers~\cite{Chang-IST2011} which is learned from the appearance of the previous $n$ frames with sampling step size $\delta$, where $n$ and $\delta$ are set as $8$ and $2$, respectively. Note that our multiple binary SVM classifiers are implemented for history reference with several unary instances, \textit{e.g.}, saliency~\cite{Liu-CVPR2016}, CNN features~\cite{Krizhevsky-NIPS2012}, location of the bounding box, and color, to segment each instance within its tracked bounding box in each frame. We only update the SVM model if the size of one instance significantly changes. We then utilize GrabCut~\cite{Rother-SIGGRAPH2004} for each instance to separate it from the background. After this step, each pixel is assigned with the instance ID.

Specifically for human instance, in case the instance is missing and re-appears in the next couple of frames, we adopt the state-of-the-art image parser,  Pyramid Scene Parsing (PSPNet)~\cite{Zhao-CVPR2017} with the pre-trained model on PASCAL VOC dataset~\cite{Everingham-ICCV2010}. The re-identification results from PSPNet are blended into our segmentation outcomes. 

\subsubsection{Contextual Property Extraction}

This component aims to determine the context of an instance so that we can apply an appropriate segmentation scheme for that instance. The context can be any observable properties that may affect the strategy to extract the mask of an instance in frames efficiently. In this work, we consider the following three attributes of an instance as its context: human or non-human, known or unknown category, rigid or deformable. 

The category of an instance, such as person, car, dog, etc., can be directly inferred from its initial mask using pre-trained Mask R-CNN~\cite{Kaiming-ICCV2017} on the MS-COCO dataset. 

To evaluate if an instance is rigid or deformable, we analyze the preview sequence of instance masks in the first $n_{Preview}$ frames. If there exists a homography matrix to transform the instance from the first frame to another frame for most frames in the first $n_{Preview}$ frames, we consider the instance to be rigid.

\subsection{\textbf{Contextual Segmentation}}
\label{sec:ProposedMethod3}

Each instance is segmented in different appropriate ways in this contextual segmentation, adapting to its extracted contextual properties (i.e., human/non-human, rigid/deformable, known/unknown category).

\subsubsection{Human Instance Segmentation}
\label{sec:HumanInstanceSegmentation}

We employ Mask R-CNN~\cite{Kaiming-ICCV2017}, pre-trained on the MS-COCO dataset ~\cite{Lin-ECCV2014}, to extract human segments. However, the results of Mask R-CNN may be affected by occlusion or unusual human pose.

\begin{figure}[t]
    \centering
    \includegraphics[width=1\columnwidth]{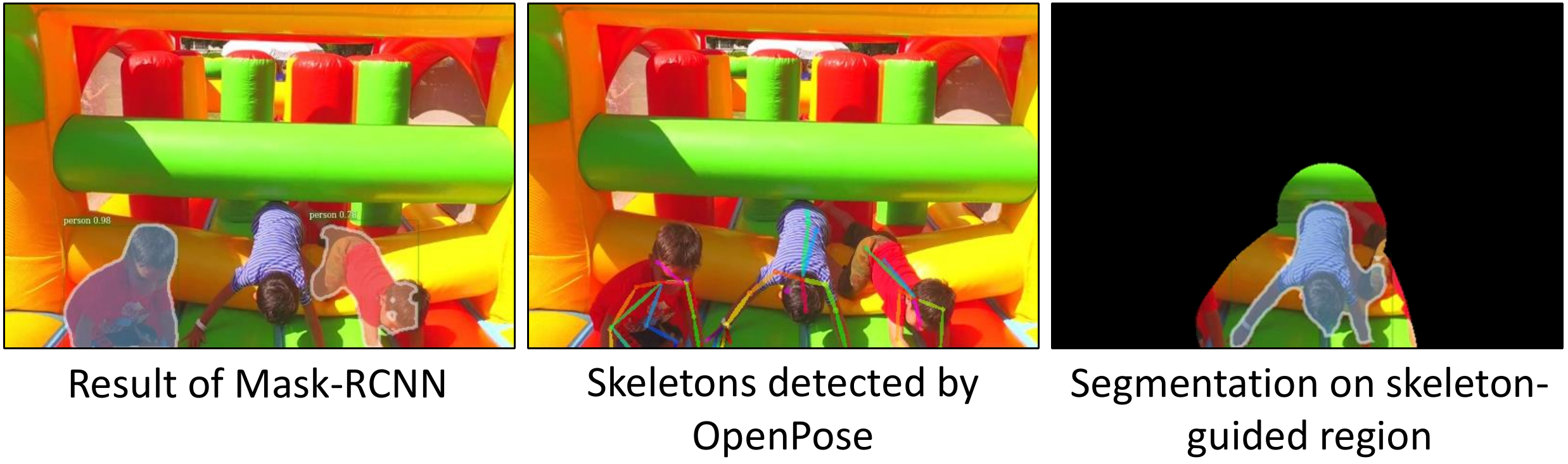}
    \caption{Skeleton-guided segmentation for unusual pose.}
    \label{fig:SkeletonGuidedMaskRCNN}
\end{figure}

\highlight{To overcome this issue, we develop \textit{skeleton-guided segmentation}. We use the skeletons from OpenPose~\cite{Cao-CVPR2017} for reference to control and refine human instance segmentation. For a human instance with an unusual pose that Mask R-CNN cannot recognize, we dilate the skeleton to obtain a skeleton-guided region, \ie an image with only the region containing the complete human instance. We then apply Mask R-CNN on a skeleton-guided region. By eliminating unrelated content, Mask R-CNN has a higher chance to extract human instance segment correctly (see Fig.~\ref{fig:SkeletonGuidedMaskRCNN}). To preserve the inter-frame mask consistency, we use object flow~\cite{Tsai-CVPR2016} to correct and refine the result across frames. }

\subsubsection{Rigid Non-Human Instance Segmentation}
\label{sec:RigidNonHumanInstanceSegmentation}

For this type of instance, our objective is to accurately extract such instances from different backgrounds in the same scene category with the initial frame. Our method to process each instance is as follows. First, we synthesize images from the first frame of a video sequence, resulting in Wonderland Data. Second, to segment instances inside bounding boxes, we train DeepLab2~\cite{Chen-TPAMI2018} and OSVOS~\cite{Caelles-CVPR2017} on our synthesized Wonderland Data.

\textbf{Wonderland Data Generation: } Differently from existing work, we exploit various contextual properties from instances. After that, multiple segmentation schemes are performed for each instance, adapting to its extracted contextual properties. Inspire by Lucid Data~\cite{Khoreva-DAVIS2017}, we introduce new augmented data, namely Wonderland Data. \highlight{To generate visual variations of the initial mask, we apply both affine and non-rigid deformations, together with illumination changes, on the mask. We also replace the background with most similar scenes filtered out from a large-scale Places365 dataset~\cite{Zhou-TPAMI2017} to preserve the semantics of the image.} In this way, we can increase more training samples than Lucid Data (10,000 images for each video, in comparing with 2,500 images of Lucid Data) to deal with one-shot learning.

\begin{figure}[t]
    \centering
    \includegraphics[width=\textwidth]{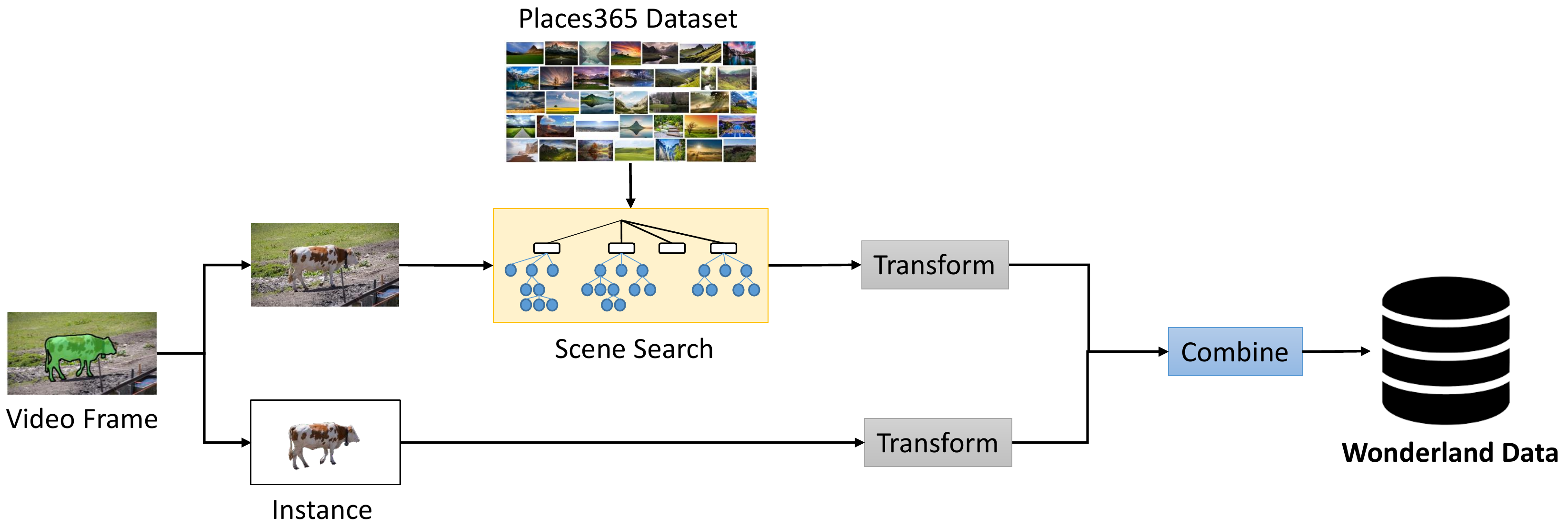}
    \caption{Wonderland Data generation.}
    \label{fig:wonderland_data}
\end{figure}

\begin{figure}[t]
    \centering
    \includegraphics[width=\textwidth]{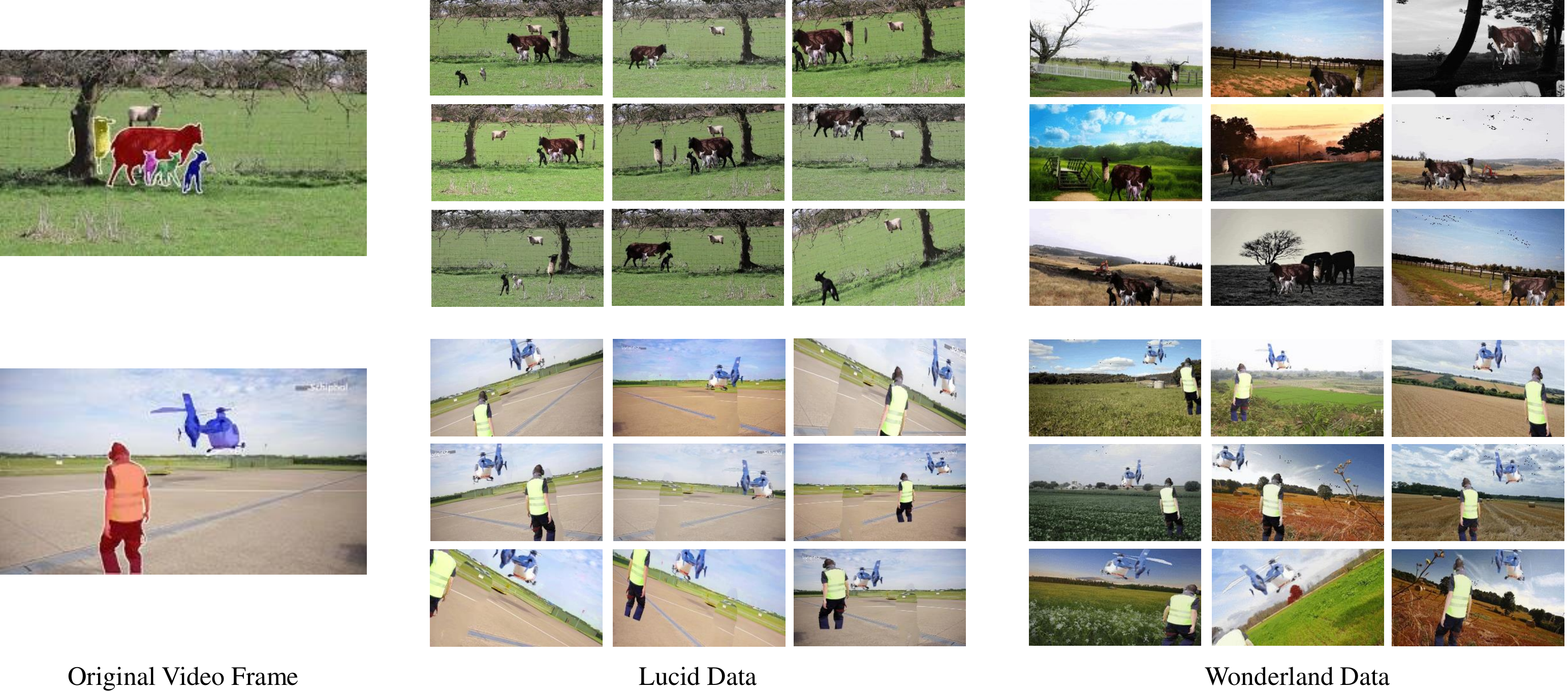}
    \caption{Augmented data generated by different methods. From left to right: the original video frames with overlaid ground-truth, followed by corresponding Lucid Data~\cite{Khoreva-DAVIS2017} and our proposed Wonderland Data in this order.}
    \label{fig:augmented_data}
\end{figure}

% \begin{figure}[t]
%     \begin{center}
%  \includegraphics[width=0.75\textwidth]{images/refinement.pdf}
%         \end{center}
%         \centering
%     \caption{Segmentation refinement based on rare instance verification.}
%     \label{fig:refinement}
% \end{figure}

Figure \ref{fig:wonderland_data} illustrates our proposed Wonderland Data generation. In this work, from a pair of an input image and a mask, we generate $10,000$ different pairs of synthesized images and masks. The Wonderland Data is published on our website\footnote{\scriptsize \url{https://sites.google.com/view/ltnghia/research/vos}}. We collect scene photos from the training set of the Places365 dataset~\cite{Zhou-TPAMI2017}, which has about 8 million images divided into 365 scene categories. We manually discard artificial scenes, use only 22 natural scene categories with 592k images. For each image, we extract a feature at the last layer of DenseNet-161~\cite{Huang-CVPR2017}, which was pre-trained on the Places365 dataset~\cite{Zhou-TPAMI2017}. This feature is used to build a hierarchical k-mean search for each category independently. We assume that each node has M images, and a leaf node has maximum $L$ images. To cluster images at a node, we propose to use $K$-mean algorithm with $K=\min(M \backslash L, T)$. In this work, we empirically set $L=200$ and $T=200$ to speed up clustering.

We classify an input image into the corresponding category, using the pre-trained DenseNet-161 on the Places365 challenge dataset. We also extract a channel feature at the last layer of the same network. After that, we search leaf nodes by comparing the Euclidean distance between the feature of an input image and the center of clusters. To search $N$ images, we randomly choose $80\%$ number of images of the nearest leaf node and $70\%, 60\%, 50\%$, etc. number of images of next leaf nodes, respectively.

We also extract the object mask from the input image, then transform the object and searched scenes independently, similarly to \cite{Khoreva-DAVIS2017}. In more detail, we use affine transformation (\eg, translation, rotation, and scale) and non-rigid deformations, together with illumination changes. Figure \ref{fig:augmented_data} shows examples of Lucid Data and our Wonderland Data.

\begin{figure}[t]
    \begin{center}
 \includegraphics[width=\textwidth]{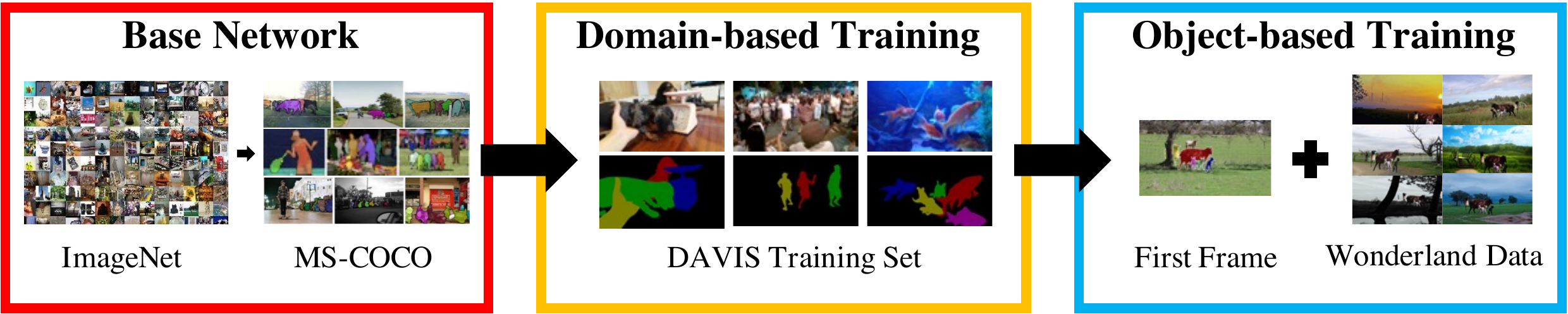}
        \end{center}
        \centering
    \caption{\highlight{The flowchart of our network training process.}}
    \label{fig:training}
\end{figure}

\textbf{Network Training:} \highlight{Figure~\ref{fig:training} our training process, including domain-based training and object-based training. In \textit{domain-based training}, we fine-tune pre-trained networks (\ie~ DeepLab2~\cite{Chen-TPAMI2018} pre-trained on COCO-Stuff dataset~\cite{Caesar-CVPR2018} and OSVOS~\cite{Caelles-CVPR2017} pre-trained on ImageNet dataset~\cite{Russakovsky-IJCV2015}) on the DAVIS training data for domain transformation. In \textit{object-based training}: we fine-tune networks on the ground-truth mask of each instance of each video. We remark that we use only the first frame of videos and apply the proposed Wonderland Data generation method for these images. }

\subsubsection{Deformable Non-Human Instance Segmentation}
\label{sec:DeformableNonHumanInstanceSegmentationInKnownUnknownCategories}

For this instance type, we categorize instances into two groups, namely, known and unknown categories. For the known categories, \textit{i.e.}, already listed in MS-COCO dataset~\cite{Lin-ECCV2014}, we simply adopt Mask R-CNN to retrieve the instance segments. We directly obtain the preview results from our IRIF component for the unknown categories since it can handle arbitrary object categories.

\begin{figure}[!t]
    \begin{center}
  \includegraphics[width=\linewidth]{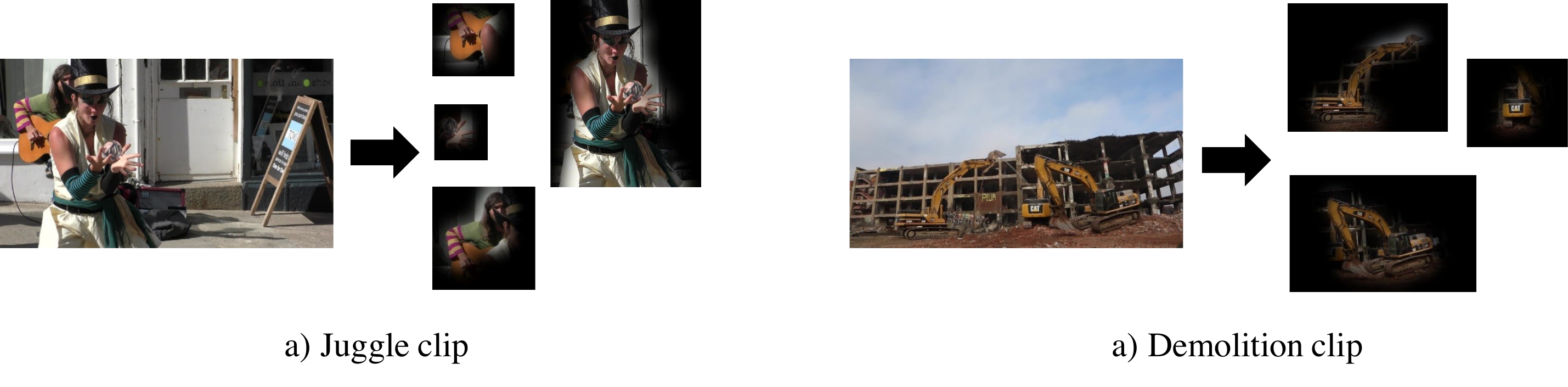}
        \end{center}
        \centering
    \caption{\highlight{Visualization of guided non-rectangular ROI.}}
    \label{fig:non-rectangular}
\end{figure}

\subsection{\textbf{Guided Segmentation}}
\label{sec:guided}

\highlight{Traditional Fully Convolutional Networks (FCNs) consider the entire rectangular region of interest (ROI) as the input to segment objects inside the ROI. This can lead to incorrect boundary segmentation due to the complex background and concave hull of the object. To overcome this limitation, we aim to transform a rectangular ROI to a non-rectangular ROI across the object boundary to eliminate the complex background inside the ROI (see Fig.~\ref{fig:non-rectangular}). In particular, we utilize referral information from extra frames to identify the shape of the instance of interest inside the ROI of the current frame. We propose to apply guided attention to construct the non-rectangular ROI and then perform fine-grained segmentation on this guided non-rectangular ROI.}

\begin{figure}[!t]
    \begin{center}
  \includegraphics[width=\linewidth]{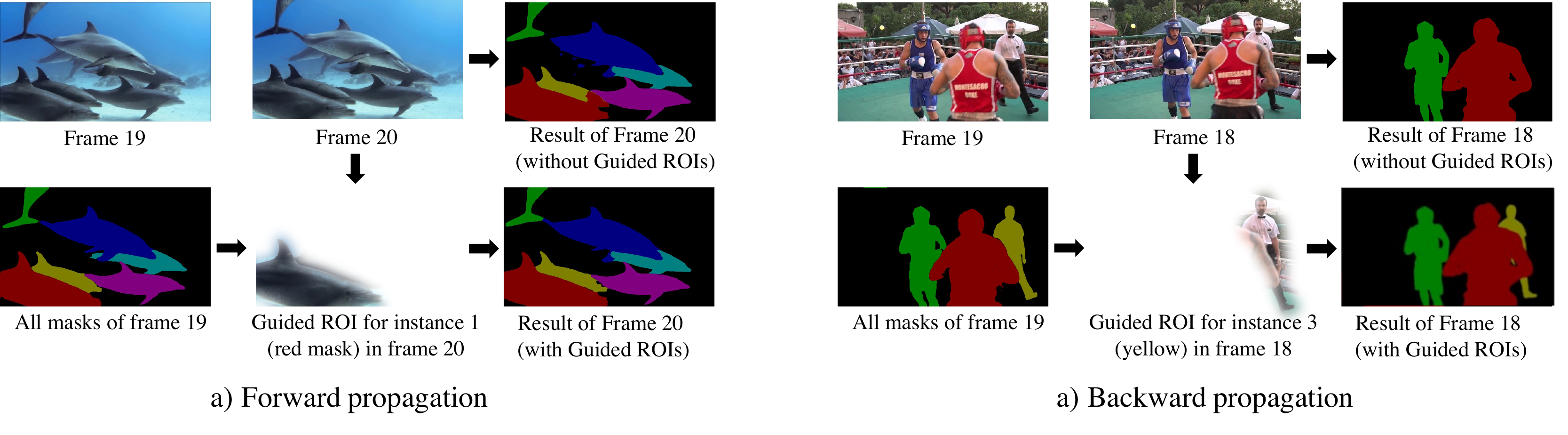}
        \end{center}
        \centering
    \caption{\highlight{Visualization of forward and backward propagation.}}
    \label{fig:propagation}
\end{figure}

\subsubsection{Bi-directional Propagation}

\highlight{In particular, we propose bi-directional strategies to construct adaptive attention for guided segmentation. Particularly, initial segments from neighbor frames are used as references for segmentation at the current frame. Attention is computed in two strategies sequentially, \textit{i.e.}, forward propagation and back-propagation, in specific ways adapting the context. Forward propagation strategy, where attention is referenced from initial segments of previous frames, can correct excessed segmentation due to dense objects in a ROI (cf. Fig.~\ref{fig:propagation}a). Meanwhile, the back-propagation strategy, where attention is referenced from initial segments of next frames, can recover missing instances due to fast motion, occlusion, or heavy deformation (size changing from tiny to large or vice versa) (cf. Fig.~\ref{fig:propagation}b).}

\subsubsection{Guided Non-Rectangular ROI Construction}

\highlight{To construct a guided non-rectangular ROI, we expand the mask of the interest instance at neighbor frames and then transfer and combine them at the current frame. This guarantees that the ROI can cover the entire interest instance. We do not apply mask propagation to avoid inaccurate flow warping as well as reducing the complexity of computation. Then, we create a smooth transition region (by applying a blurred mask to remove background) for the guided ROI to avoid a clear border between the ROI and background. It is essential to make the segmentation method focus on the interest instance and avoid inaccurate segmentation due to a clear border. We remark that the range of boundary expansion and transition smooth is computed based on the intensity of movement of the instance. Both propagation strategies are performed adaptively if initial segments of the interest instance at the current frame are much different (in appearance or size) from those at neighbor frames or the instance re-appears. On the other hand, we only refine the interest instance at the current frame to save the computational cost. }

% \begin{figure}[!t]
%     \begin{center}
%   \includegraphics[width=0.8\linewidth]{images/ForwardGuidedROI_Transparent.pdf}
%         \end{center}
%         \centering
%     \caption{\highlight{Visualization of forward propagation.}}
%     \label{fig:forward_visualization}
% \end{figure}

% \begin{figure}[!t]
%     \begin{center}
%   \includegraphics[width=0.8\linewidth]{images/BackwardGuidedROI_Transparent.pdf}
%         \end{center}
%         \centering
%     \caption{\highlight{Visualization of back-propagation.}}
%     \label{fig:backward_visualization}
% \end{figure}

\subsubsection{Fine-grained Segmentation}

\highlight{We use Deep Grabcut~\cite{Xu-BMVC2017} and Mask R-CNN~\cite{Kaiming-ICCV2017} for fine-grained segmentation in guided non-rectangular ROIs. Inspired by Luiten et.al.~\cite{Luiten-DAVIS2018}, we train DeepLab3+~\cite{Chen-ECCV2018} based on Xception-65~\cite{Chollet-CVPR2017} backbone on MS-COCO~\cite{Lin-ECCV2014} and Mapillary~\cite{Neuhold-ICCV2017} datasets to enhance the network generalization. For Mask R-CNN, we directly use a pre-trained model on MS-COCO~\cite{Lin-ECCV2014} dataset.}

\subsection{\textbf{Refinement and Merging}}
\label{sec:refinement}

Through preliminary results, we observe that the initial segmentation is not smooth enough. Therefore, we refine instance masks to improve segmentation quality, using rare instance attention and boundary snapping.

\subsubsection{Rare-Instance Attention Refinement}

%Figure~\ref{fig:refinement} illustrates our proposed segmentation refinement based on rare instance attention. 

We further refine the results by considering the rare instances. We observe that rare objects are shrunk due to larger objects. To identify rare object instances, we compute each object instance mask percentage in terms of area (provided in the first frame). Instances with a size smaller than 5\% the total size of tracking objects are considered rare ones. We assume that a smaller object tends to be small in the whole video. Next, we recover rare object instances by transferring the results produced by the foreground probability obtained from the binary-SVM classifier on each object instance.

\subsubsection{Boundary Snapping Refinement}

We also adopt boundary snapping~\cite{Caelles-CVPR2017} to further refine object shapes. In particular, we extract the saliency~\cite{Liu-CVPR2016} and the contour~\cite{Yang-CVPR2016} from the video frame. The salient pixels close to the contour are snapped.  

\subsubsection{Topological Order Estimation for Instance Merging}
\label{sec:InstanceMerging}

It is essential to determine the topology relationship (in terms of $z$-order) between multiple instances to sequentially combine corresponding masks of different instances into the final result. We here merge instances based on human and non-human instance interaction, depth values, and rare instance priority heuristics in this order as follows:

\begin{itemize}

\item \textbf{Human and non-human instance interaction}: We define interaction heuristics as follow: transportation instances (such as horse, bike, motor, surfboard, and skateboard, etc.) are the farthest from the camera; human instance have the middle distance to the camera; and small non-human instances which can be held, bring, touch, etc. are the nearest from the camera. Interacted small non-human instances are localized at the human hand's position using OpenPose~\cite{Cao-CVPR2017}.

\item \textbf{Depth values}: We first estimate pixel-wise depth values of the video frame, using DCNF-FCSP~\cite{Liu-CVPR2015}, and then take the average value for each instance.

\item \textbf{Rare instance priority}: We notice that rare instances are always the nearest ones from the camera.

\end{itemize}

%-------------------------------------------------------------------------

\section{Experimental Results}
\label{sec:results}

\subsection{\textbf{Dataset Benchmark and Metrics}}

We participated the DAVIS Challenges 2017-2019, Semi-Supervised Track\footnote{\url{https://davischallenge.org/challenge2017/index.html}}\footnote{\url{https://davischallenge.org/challenge2018/index.html}}\footnote{\url{https://davischallenge.org/challenge2019/index.html}} and evaluated our methods on the \textit{DAVIS Test-Challenge} dataset. The dataset consists of $150$ sequences, totaling $10,459$ annotated frames and $376$ instances. There are a total of 30 video sequences for testing, and their ground truth not publicly available. Submissions were made through the CodaLab site of the challenge\footnote{\url{https://competitions.codalab.org/competitions/21650}}. This dataset is challenging due to multiple object instances with more distractors, \textit{i.e.}, smaller instances and fine structures, more occlusions, and fast motion.

For the evaluation metrics, per-instance measures are used as described in~\cite{Perazzi-CVPR2016}: Region Jaccard (J) and Boundary F measure (F). The overall measures are computed as the mean between J and F, and both are averaged over all objects.

\begin{table}[t]
\centering
\caption{\highlight{Top global ranking results in the DAVIS Challenges 2017-2019. The best results are marked in \textbf{boldface}. Our results are marked in \textcolor[rgb]{0,0,1}{blue}. We note that the teams without references do not have publication.} }
\resizebox{\textwidth}{!}{%
\begin{tabular}{clc|ccccccccc}
\hline 
\multicolumn{1}{c}{\multirow{2}{*}{\textbf{Rank}}} & \multicolumn{1}{c}{\multirow{2}{*}{\textbf{Method/Team}}} &
\multicolumn{1}{c|}{\multirow{2}{*}{\textbf{Year}}} & \multicolumn{1}{c}{\textbf{Global G}} & \textbf{} & \multicolumn{3}{c}{\textbf{Region J}} & \textbf{} & \multicolumn{3}{c}{\textbf{Boundary F}} \\ \cline{4-4} \cline{6-8} \cline{10-12} 
\multicolumn{1}{c}{} & \multicolumn{1}{c}{} & \multicolumn{1}{c|}{} & \multicolumn{1}{c}{\textbf{Mean $\Uparrow$}} & \textbf{} & \multicolumn{1}{c}{\textbf{Mean $\Uparrow$}} & \multicolumn{1}{c}{\textbf{Recall $\Uparrow$}} & \multicolumn{1}{c}{\textbf{Decay $\Downarrow$}} & \textbf{} & \multicolumn{1}{c}{\textbf{Mean $\Uparrow$}} & \multicolumn{1}{l}{\textbf{Recall $\Uparrow$}} & \multicolumn{1}{c}{\textbf{Decay $\Downarrow$}} \\ \hline

1 & OSS~\cite{Wang-DAVIS2019} & 2019 & \textbf{76.7} & & 72.8 & 81.5 & 18.9 & & \textbf{80.7} & 87.5 & 21.3 \\
2 & BoLTVOS+~\cite{Luiten-DAVIS2019} & 2019 & 76.2 & & \textbf{72.9} & \textbf{81.7} & 16.3 & & 79.4 & 86.7 & 19.5 \\
\textcolor[rgb]{0,0,1}{3} & \textcolor[rgb]{0,0,1}{CGS~\cite{tmtriet-DAVIS2019}} & \textcolor[rgb]{0,0,1}{2019} & \textcolor[rgb]{0,0,1}{75.4} & & \textcolor[rgb]{0,0,1}{72.4} & \textcolor[rgb]{0,0,1}{\textbf{81.7}} & \textcolor[rgb]{0,0,1}{\textbf{11.0}} & & \textcolor[rgb]{0,0,1}{78.4} & \textcolor[rgb]{0,0,1}{\textbf{87.6}} & \textcolor[rgb]{0,0,1}{\textbf{12.9}} \\ 
4 & STM~\cite{Oh-DAVIS2019} & 2019 & 75.2 & & 72.6 & 80.9 & 21.0 & & 77.7 & 85.0 & 24.1 \\
5 & PremVOS~\cite{Luiten-DAVIS2018} & 2018 & 74.7 & & 71.0 & 79.5 & 19.0 & & 78.4 & 86.7 & 20.8 \\ 
6 & DyeNet~\cite{Li-DAVIS2018} & 2018 & 73.8 & & 71.9 & 79.4 & 19.8 & & 75.8 & 83.0 & 20.3 \\
7 & Theodoruszq & 2019 & 73.1 & & 70.1 & 77.3 & 24.8 & & 76.1 & 84.0 & 28.3 \\
8 & Panday & 2019 & 71.3 & & 67.7 & 74.8 & 24.7 & & 75.0 & 81.2 & 27.5 \\
9 & DLTA~\cite{Robinson-DAVIS2019} & 2019 & 70.6 & & 68.5 & 78.1 & 20.3 & & 72.8 & 84.2 & 24.0 \\
10 & VS-ReID~\cite{Li-DAVIS2017} & 2017 & 69.9 & & 67.9 & 74.6 & 25.5 & & 71.9 & 79.1 & 24.1 \\
11 & CAVOS~\cite{Xu-DAVIS2018} & 2018 & 69.7 & & 66.9 & 74.1 & 23.1 & & 72.5 & 80.3 & 25.9 \\
12 & ODG~\cite{Guo-DAVIS2018} & 2018 & 69.5 & & 67.5 & 77.0 & 15.0 & & 71.5 & 82.2 & 18.5 \\
13 & PVOS~\cite{Guo-DAVIS2019} & 2019 & 69.2 & & 66.0 & 73.4 & 28.5 & & 72.3 & 80.4 & 31.1 \\
14 & LucidTracker~\cite{Khoreva-DAVIS2017} & 2017 & 67.8 & & 65.1 & 72.5 & 27.7 & & 70.6 & 79.8 & 30.2 \\
\textcolor[rgb]{0,0,1}{15} & \textcolor[rgb]{0,0,1}{Second Pass~\cite{tmtriet-DAVIS2018}} & \textcolor[rgb]{0,0,1}{2018} & \textcolor[rgb]{0,0,1}{66.3} & & \textcolor[rgb]{0,0,1}{64.1} & \textcolor[rgb]{0,0,1}{75.0} & \textcolor[rgb]{0,0,1}{11.7} & & \textcolor[rgb]{0,0,1}{68.6} & \textcolor[rgb]{0,0,1}{80.7} & \textcolor[rgb]{0,0,1}{13.5} \\
\textcolor[rgb]{0,0,1}{16} & \textcolor[rgb]{0,0,1}{First Pass~\cite{ltnghia-DAVIS2017}} & \textcolor[rgb]{0,0,1}{2017} & \textcolor[rgb]{0,0,1}{63.8} & & \textcolor[rgb]{0,0,1}{61.5} & \textcolor[rgb]{0,0,1}{68.6} & \textcolor[rgb]{0,0,1}{17.1} & & \textcolor[rgb]{0,0,1}{66.2} & \textcolor[rgb]{0,0,1}{79.0} & \textcolor[rgb]{0,0,1}{17.6} \\
17 & SPT~\cite{Shaban-DAVIS2017} & 2017 & 61.5 & & 59.8 & 71.0 & 21.9 & & 74.6 & 74.6 & 23.7 \\
18 & FAVOS~\cite{Lin-DAVIS2018} & 2018 & 60.6 & & 58.4 & 65.6 & 26.2 & & 62.9 & 71.0 & 29.7 \\
19 & MPN~\cite{Sun-DAVIS2018} & 2018 & 60.1 & & 57.7 & 64.9 & 27.2 & & 62.4 & 71.7 & 28.1 \\
20 & PALC~\cite{Petrosyan-DAVIS2018} & 2018 & 58.9 & & 56.7 & 63.1 & 30.7 & & 61.1 & 67.6 & 33.1 \\
21 & OnAVOS~\cite{Voigtlaender-DAVIS2017} & 2017 & 57.7 & & 54.8 & 60.8 & 60.5 & & 67.2 & 67.2 & 34.7 \\ 
22 & SPN~\cite{Cheng-DAVIS2017} & 2017 & 56.9 & & 54.8 & 60.7 & 34.4 & & 59.1 & 66.7 & 36.1 \\
23  & HE-PSPNet~\cite{Zhao-DAVIS2017} & 2017  & 56.9 &  & 53.6 & 59.5 & 25.3 &  & 60.2 & 67.9 & 27.6 \\
24  & OSVOS-IOFT~\cite{Newswanger-DAVIS2017} & 2017  & 55.8 &  & 53.8 & 60.1 & 37.7 &  & 57.8 & 62.1 & 42.9 \\
25  & TOP~\cite{Sharir-DAVIS2017} & 2017  & 54.8 &  & 51.6 & 56.3  & 26.8 &  & 57.9 & 64.8 & 28.8 \\
26 & Froma & 2017  & 53.9 &  & 50.9 & 54.9 & 32.5 &  & 57.1 & 66.2 & 33.7 \\
 \hline 
\end{tabular}
}
\label{tab:results1}
\end{table}

\begin{figure}[t]
    \begin{center}
  \includegraphics[width=\textwidth]{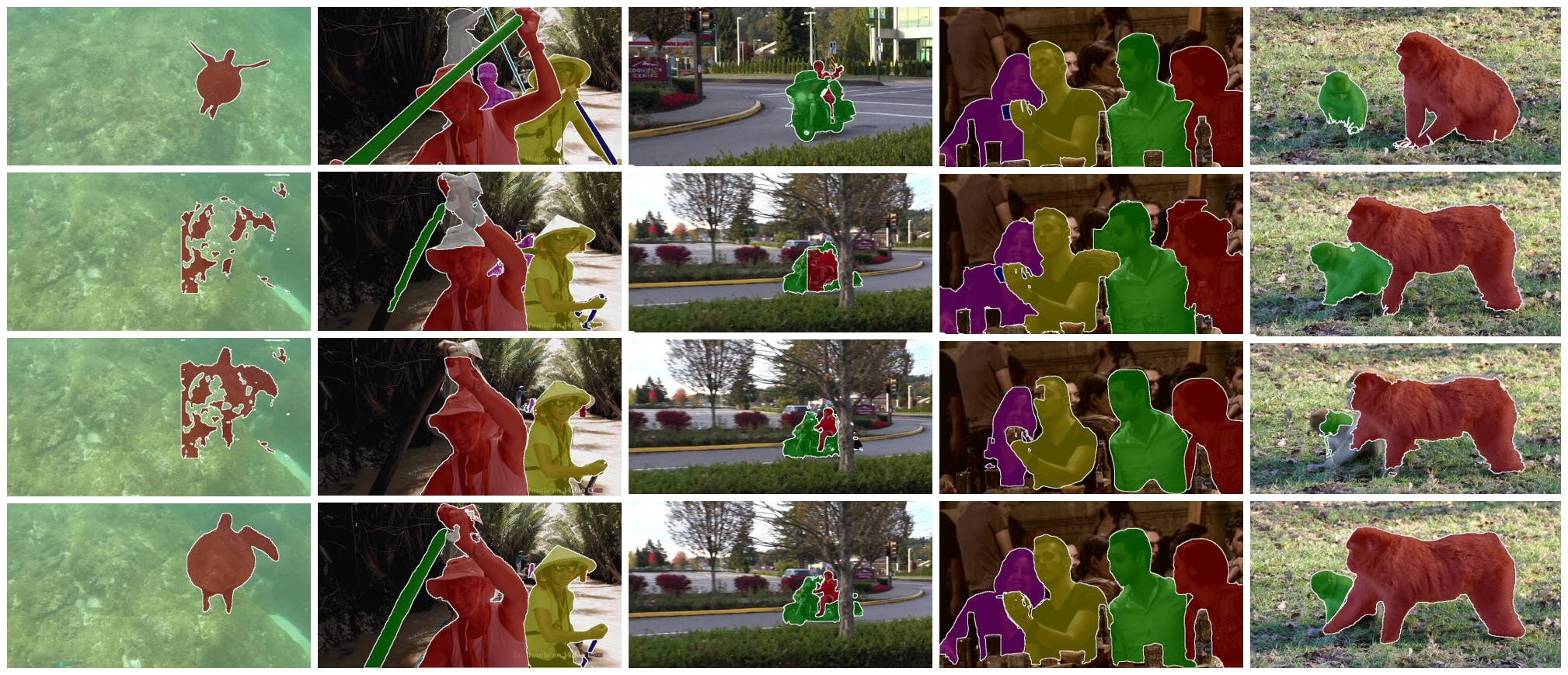}
        \end{center}
        \centering
    \caption{\highlight{Visualization results on the DAVIS Test-Challenge dataset. From top to bottom: the first video frame with the ground-truth label followed by results of our proposed methods in preview segmentation~\cite{ltnghia-DAVIS2017}, contextual segmentation~\cite{tmtriet-DAVIS2018}, and guided segmentation~\cite{tmtriet-DAVIS2019}. The ground-truth of the certain video frame is not publicly available. Our CGS results significantly track and segment the instances of interest as annotated in the first frame.}}
    
    \label{fig:visualization}
\end{figure}

\subsection{\textbf{Results on DAVIS Challenges 2017-2019}}

\subsubsection{DAVIS 2017 Challenge}

Due to the time limit, we submitted the proposed IRIF component in the DAVIS 2017 Challenge and achieved $3^{rd}$ place out of 22 team submissions in this challenge. As shown in Table~\ref{tab:results1}, our proposed IRIF achieves very promising results in the DAVIS 2017 Challenge, namely, 0.615, 0.662, and 0.638 in terms of region similarity (Jaccard index), contour accuracy (F-measure), and global score, respectively. Our results highly indicate that our method is competitive among the state-of-the-art methods in this dataset. Our method maintains the performance as frames evolve, as seen via the best performance in terms of J decay and F decay among the leading submissions in 2017.

\subsubsection{DAVIS 2018 Challenge}

\highlight{We also had another submission of CIS framework to the DAVIS 2018 Challenge and achieved $6^{th}$ place out of 41 team submissions in this challenge. Table~\ref{tab:results1} shows that our CIS achieves promising results, namely, 64.1\%, 68.6\%, and 66.3\% in terms of region similarity (Jaccard index), contour accuracy (F-measure), and global score, respectively. Our method also maintains the best stable performance in terms of J decay and F decay among the leading submissions in 2018. }

\subsubsection{DAVIS 2019 Challenge}

\highlight{As shown in Table~\ref{tab:results1}, we obtained very competitive results. Our proposed CGS achieved 0.724, 0.784, and 0.754 in terms of region similarity (J), contour accuracy (F), and global score, respectively. Our method achieved the best performance in Decay and Recall of all metrics consistently. Furthermore, we note that our CGS is in top 3 over 4 teams achieving 0.75 in terms of global score in all three years. }

\subsubsection{Ablation Study}

\highlight{Table~\ref{tab:results2} shows the results of our proposed framework with different settings. Our proposed CGS (using all three passes) outperforms using only two passes~\cite{tmtriet-DAVIS2018} or a pass~\cite{ltnghia-DAVIS2017}. This highlights the significant contribution of the second pass and the third pass, which are the multiple contextual segmentation schemes, and guided instance segmentation, respectively. Particularly, contextual segmentation can improve the performance up to 2.5\%. Meanwhile, guided segmentation improves contextual segmentation up to 9.1\% in the global score.}

\highlight{Figure~\ref{fig:visualization} visualizes segmentation results. From top row to bottom row, we can observe the first video frame and a triple of processed video frames of our proposed methods in preview segmentation~\cite{ltnghia-DAVIS2017}, contextual segmentation~\cite{tmtriet-DAVIS2018}, and guided segmentation~\cite{tmtriet-DAVIS2019}. Our final CGS results surpass the performance of others and successfully track and segment the key instances. Our framework can even handle camouflaged instances, small instances, and occluded instances.}

\begin{table*}[t]
\centering
\caption{\highlight{The performance of different components in our method on the DAVIS Test-Challenge dataset. PS, CS, and GS stand for preview segmentation, contextual segmentation, and guided segmentation, respectively.}}
\footnotesize
%\resizebox{\textwidth}{!}{%
\begin{tabular}{ccc|ccc}
\hline \hline

\multicolumn{3}{c|}{\textbf{Settings}} & \textbf{Global Score} $\Uparrow$ & \textbf{Region J} $\Uparrow$ & \textbf{Boundary F} $\Uparrow$ \\
\cline{1-3}
\textbf{PS} & \textbf{CS} & \textbf{GS} & & & \\
\hline
\cmark & &  & 63.8 & 61.5 & 66.2 \\
\cmark & \cmark & &  66.3 & 64.1 & 68.6 \\
\cmark & \cmark & \cmark & \textbf{75.4} & \textbf{72.4} & \textbf{78.4} \\

 \hline \hline
\end{tabular}
%}
\label{tab:results2}
\end{table*}

%-------------------------------------------------------------------------
\section{Conclusion}
\label{sec:conclusion}

\highlight{In this paper, we propose the novel CGS framework for semi-supervised instance segmentation in videos with three segmentation passes. In the first pass, we develop the novel IRIF for preview instance segmentation and extract contextual information. In the second pass, we introduce multiple contextual segmentation schemes to deal with different instance types, such as human/non-human rigid/non-rigid instances in known/unknown object categories. In the final pass, we propose a novel guided fined-grained segmentation based on attention to eliminate complex background inside the region of interest for performance improvement.}

\highlight{Our proposed methods achieve competitive results among the leading submissions in the DAVIS Challenges consistently, \ie \textbf{$3^{rd}$} place, \textbf{$6^{th}$} place, and \textbf{$3^{rd}$} place in 2017, 2018, and 2019, respectively. Our full framework CGS is in the top 3 over 4 teams achieving 0.75 in terms of global score in all three years. Our method also maintains the best stable and recall performance among the leading submissions.}

In the future, we plan to consider modeling the semantic relationship among object instances in the segmentation process. \highlight{We will also investigate Capsule-inspired~\cite{Hinton-ICANN2011, Sabour-2020, Sabour-NeurIPS2017, Zhang-RS2019}, and attention-inspired~\cite{ltnghia-IV2020, Dosovitskiy-ICLR2021, Brendan-CVPR2021, Carion-ECCV2020} network architectures for better segmentation performance. We also aim to extend our work to camouflage analysis~\cite{ltnghia-CVIU2019, Jinnan-IEEEAccess2021, ltnghia-2021} in the near future. }

%-------------------------------------------------------------------------

\section*{Acknowledgment}

This research is funded by Gia Lam Urban Development and Investment Company Limited, Vingroup, supported by Vingroup Innovation Foundation (VINIF) under project code VINIF.2019.DA19, and National Science Foundation (NSF) under Grant No. 2025234. The first author would like to thank JSPS KAKENHI Grants (JP16H06302, JP18H04120, JP21H04907, JP20K23355, JP21K18023), JST CREST Grants (JPMJCR20D3, JPMJCR18A6). We also thank NVIDIA and AIOZ Pte Ltd for the support of GPU and computing infrastructure.

%\clearpage

\balance
{\normalsize
\bibliographystyle{ieee}
\bibliography{shortbib}

\begin{thebibliography}{10}\itemsep=-1pt

\bibitem{Brabandere-CVPRW2017}
B.~D. {Brabandere}, D.~{Neven}, and L.~V. {Gool}.
\newblock Semantic instance segmentation for autonomous driving.
\newblock In {\em CVPR Workshops}, 2017.

\bibitem{Caelles-CVPR2017}
S.~Caelles, K.-K. Maninis, J.~Pont-Tuset, L.~Leal-Taix\'e, D.~Cremers, and
  L.~{Van Gool}.
\newblock One-shot video object segmentation.
\newblock In {\em CVPR}, 2017.

\bibitem{Caesar-CVPR2018}
H.~Caesar, J.~Uijlings, and V.~Ferrari.
\newblock Coco-stuff: Thing and stuff classes in context.
\newblock In {\em CVPR}, 2018.

\bibitem{Cao-CVPR2017}
Z.~Cao, T.~Simon, S.-E. Wei, and Y.~Sheikh.
\newblock Realtime multi-person 2d pose estimation using part affinity fields.
\newblock In {\em CVPR}, 2017.

\bibitem{Carion-ECCV2020}
N.~Carion, F.~Massa, G.~Synnaeve, N.~Usunier, A.~Kirillov, and S.~Zagoruyko.
\newblock End-to-end object detection with transformers.
\newblock In {\em ECCV}, pages 213--229, 2020.

\bibitem{Chang-IST2011}
C.~Chang and C.~Lin.
\newblock {LIBSVM:} {A} library for support vector machines.
\newblock {\em {Transactions on Intelligent Systems and Technology }}, 2(3),
  2011.

\bibitem{Chen-TPAMI2018}
L.~C. Chen, G.~Papandreou, I.~Kokkinos, K.~Murphy, and A.~L. Yuille.
\newblock Deeplab: Semantic image segmentation with deep convolutional nets,
  atrous convolution, and fully connected crfs.
\newblock {\em Transactions on Pattern Analysis and Machine Intelligence},
  40(4), 2018.

\bibitem{Chen-ECCV2018}
L.-C. Chen, Y.~Zhu, G.~Papandreou, F.~Schroff, and H.~Adam.
\newblock Encoder-decoder with atrous separable convolution for semantic image
  segmentation.
\newblock In {\em ECCV}, 2018.

\bibitem{Cheng-DAVIS2017}
J.~Cheng, S.~Liu, Y.-H. Tsai, W.-C. Hung, S.~Gupta, J.~Gu, J.~Kautz, S.~Wang,
  and M.-H. Yang.
\newblock Learning to segment instances in videos with spatial propagation
  network.
\newblock {\em CVPR Workshops}, 2017.

\bibitem{Chollet-CVPR2017}
F.~Chollet.
\newblock Xception: Deep learning with depthwise separable convolutions.
\newblock In {\em CVPR}, 2017.

\bibitem{Dosovitskiy-ICLR2021}
A.~Dosovitskiy, L.~Beyer, A.~Kolesnikov, D.~Weissenborn, X.~Zhai,
  T.~Unterthiner, M.~Dehghani, M.~Minderer, G.~Heigold, S.~Gelly, et~al.
\newblock An image is worth 16x16 words: Transformers for image recognition at
  scale.
\newblock {\em ICLR}, 2021.

\bibitem{Brendan-CVPR2021}
B.~Duke, A.~Ahmed, C.~Wolf, P.~Aarabi, and G.~W. Taylor.
\newblock Sstvos: Sparse spatiotemporal transformers for video object
  segmentation.
\newblock In {\em CVPR}, 2021.

\bibitem{Everingham-ICCV2010}
M.~Everingham, L.~J.~V. Gool, C.~K.~I. Williams, J.~M. Winn, and A.~Zisserman.
\newblock The pascal visual object classes {(VOC)} challenge.
\newblock {\em IJCV}, 88(2), 2010.

\bibitem{Pedro-CVPR2008}
P.~F. Felzenszwalb, D.~A. McAllester, and D.~Ramanan.
\newblock A discriminatively trained, multiscale, deformable part model.
\newblock In {\em CVPR}, 2008.

\bibitem{Fiaz-DAVIS2020}
M.~Fiaz, A.~Mahmood, and S.~K. Jung.
\newblock Video object segmentation using guided feature and directional deep
  appearance learning.
\newblock {\em CVPR Workshops}, 2020.

\bibitem{Guo-DAVIS2019}
H.~Guo, W.~Wang, G.~Guo, H.~Li, J.~Liu, Q.~He, and X.~Xiao.
\newblock An empirical study of propagation-based methods for video object
  segmentation.
\newblock {\em CVPR Workshops}, 2019.

\bibitem{Guo-DAVIS2018}
P.~Guo, L.~Zhang, H.~Zhang, X.~Liu, H.~Ren, and Y.~Zhang.
\newblock Adaptive video object segmentation with online data generation.
\newblock {\em CVPR Workshops}, 2018.

\bibitem{Kaiming-ICCV2017}
K.~He, G.~Gkioxari, P.~Doll{\'a}r, and R.~Girshick.
\newblock Mask r-cnn.
\newblock In {\em ICCV}, 2017.

\bibitem{Hinton-ICANN2011}
G.~E. Hinton, A.~Krizhevsky, and S.~D. Wang.
\newblock Transforming auto-encoders.
\newblock In {\em International conference on artificial neural networks},
  pages 44--51, 2011.

\bibitem{Huang-CVPR2017}
G.~Huang, Z.~Liu, L.~van~der Maaten, and K.~Q. Weinberger.
\newblock Densely connected convolutional networks.
\newblock In {\em CVPR}, 2017.

\bibitem{Ji-ECCV2018}
J.~Ji, S.~Buch, A.~Soto, and J.~C. Niebles.
\newblock End-to-end joint semantic segmentation of actors and actions in
  video.
\newblock In {\em ECCV}, 2018.

\bibitem{Khoreva-DAVIS2017}
A.~Khoreva, R.~Benenson, E.~Ilg, T.~Brox, and B.~Schiele.
\newblock Lucid data dreaming for object tracking.
\newblock {\em CVPR Workshops}, 2017.

\bibitem{Krizhevsky-NIPS2012}
A.~Krizhevsky, I.~Sutskever, and G.~E. Hinton.
\newblock Imagenet classification with deep convolutional neural networks.
\newblock In {\em Advances in Neural Information Processing Systems}, 2012.

\bibitem{ltnghia-2021}
T.-N. Le, Y.~Cao, T.-C. Nguyen, M.-Q. Le, K.-D. Nguyen, T.-T. Do, M.-T. Tran,
  and T.~V. Nguyen.
\newblock Camouflaged instance segmentation in-the-wild: Dataset and benchmark
  suite.
\newblock {\em ArXiv Pre-print: 2103.17123}, 2021.

\bibitem{Le-DAVIS2017}
T.-N. Le, K.-T. Nguyen, M.-H. Nguyen-Phan, T.-V. Ton, T.-A. Nguyen, X.-S.
  Trinh, Q.-H. Dinh, V.-T. Nguyen, A.-D. Duong, A.~Sugimoto, T.~V. Nguyen, and
  M.-T. Tran.
\newblock Instance re-identification flow for video object segmentation.
\newblock {\em CVPR Workshops}, 2017.

\bibitem{ltnghia-DAVIS2017}
T.-N. Le, K.-T. Nguyen, M.-H. Nguyen-Phan, V.~Ton-That, T.-A. Nguyen, X.-S.
  Trinh, Q.-H. Dinh, V.-T. Nguyen, A.~D. Duong, A.~Sugimoto, T.~V. Nguyen, and
  M.-T. Tran.
\newblock Instance re-identification flow for video object segmentation.
\newblock {\em CVPR Workshops}, 2017.

\bibitem{ltnghia-CVIU2019}
T.-N. Le, T.~V. Nguyen, Z.~Nie, M.-T. Tran, and A.~Sugimoto.
\newblock Anabranch network for camouflaged object segmentation.
\newblock {\em Journal of Computer Vision and Image Understanding}, 184:45--56,
  2019.

\bibitem{ltnghia-AAAI2021}
T.-N. Le, T.~V. Nguyen, Q.-C. Tran, L.~Nguyen, T.-H. Hoang, M.-Q. Le, and M.-T.
  Tran.
\newblock Interactive video object mask annotation.
\newblock In {\em AAAI}, 2021.

\bibitem{ltnghia-IV2020}
T.-N. Le, A.~Sugimoto, S.~Ono, and H.~Kawasaki.
\newblock Attention r-cnn for accident detection.
\newblock In {\em IEEE Intelligent Vehicles Symposium}, 2020.

\bibitem{Lee-IJCV2015}
Y.~J. Lee and K.~Grauman.
\newblock Predicting important objects for egocentric video summarization.
\newblock {\em IJCV}, 114(1), 2015.

\bibitem{Li-DAVIS2018}
X.~Li and C.~C. Loy.
\newblock Video object segmentation with joint re-identification and
  attention-aware mask propagation.
\newblock {\em CVPR Workshops}, 2018.

\bibitem{Li-DAVIS2017}
X.~Li, Y.~Qi, Z.~Wang, K.~Chen, Z.~Liu, J.~Shi, P.~Luo, C.~C. Loy, and X.~Tang.
\newblock Video object segmentation with re-identification.
\newblock {\em CVPR Workshops}, 2017.

\bibitem{Lin-DAVIS2018}
A.~Lin, Y.~Chou, and T.~Martinez.
\newblock Flow adaptive video object segmentation.
\newblock {\em CVPR Workshops}, 2018.

\bibitem{Lin-ECCV2014}
T.-Y. Lin, M.~Maire, S.~Belongie, J.~Hays, P.~Perona, D.~Ramanan,
  P.~Doll{\'a}r, and C.~L. Zitnick.
\newblock Microsoft coco: Common objects in context.
\newblock In {\em ECCV}, 2014.

\bibitem{Liu-DAVIS2020}
D.~Liu, D.~Yu, M.~Dong, L.~Ma, J.~Shao, J.~Wang, C.~Wang, and P.~Zhou.
\newblock An effective multi-level backbone for video object segmentation.
\newblock {\em CVPR Workshops}, 2020.

\bibitem{Liu-CVPR2016}
N.~Liu and J.~Han.
\newblock Dhsnet: Deep hierarchical saliency network for salient object
  detection.
\newblock In {\em CVPR}, 2016.

\bibitem{Liu-CVPR2015}
N.~Liu, J.~Han, D.~Zhang, S.~Wen, and T.~Liu.
\newblock Predicting eye fixations using convolutional neural networks.
\newblock In {\em CVPR}, 2015.

\bibitem{Luiten-DAVIS2018}
J.~Luiten, P.~Voigtlaender, and B.~Leibe.
\newblock Premvos: Proposal-generation, refinement and merging for the davis
  challenge on video object segmentation.
\newblock {\em CVPR Workshops}, 2018.

\bibitem{Luiten-DAVIS2019}
J.~Luiten, P.~Voigtlaender, and B.~Leibe.
\newblock Combining premvos with box-level tracking for the 2019 davis
  challenge.
\newblock {\em CVPR Workshops}, 2019.

\bibitem{Neuhold-ICCV2017}
G.~Neuhold, T.~Ollmann, S.~Rota~Bulo, and P.~Kontschieder.
\newblock The mapillary vistas dataset for semantic understanding of street
  scenes.
\newblock In {\em ICCV}, 2017.

\bibitem{Newswanger-DAVIS2017}
A.~Newswanger and C.~Xu.
\newblock One-shot video object segmentation with iterative online fine-tuning.
\newblock {\em CVPR Workshops}, 2017.

\bibitem{YADA}
K.~Nguyen, K.~Nguyen, D.~Le, D.~A. Duong, and T.~V. Nguyen.
\newblock {YADA:} you always dream again for better object detection.
\newblock {\em Multim. Tools Appl.}, 78(19):28189--28208, 2019.

\bibitem{YALA}
K.~Nguyen, K.~Nguyen, D.~Le, D.~A. Duong, and T.~V. Nguyen.
\newblock You always look again: Learning to detect the unseen objects.
\newblock {\em J. Vis. Commun. Image Represent.}, 60:206--216, 2019.

\bibitem{Oh-DAVIS2019}
S.~W. Oh, J.~Lee, N.~Xu, and S.~J. Kim.
\newblock A unified model for semi-supervised and interactive video object
  segmentation using space-time memory networks.
\newblock {\em CVPR Workshops}, 2019.

\bibitem{Perazzi-CVPR2016}
F.~Perazzi, J.~Pont-Tuset, B.~McWilliams, L.~V. Gool, M.~Gross, and
  A.~Sorkine-Hornung.
\newblock A benchmark dataset and evaluation methodology for video object
  segmentation.
\newblock In {\em CVPR}, 2016.

\bibitem{Jordi-2017}
J.~Pont-Tuset, F.~Perazzi, S.~Caelles, P.~Arbel\'aez, A.~Sorkine-Hornung, and
  L.~{Van Gool}.
\newblock The 2017 davis challenge on video object segmentation.
\newblock {\em arXiv:1704.00675}, 2017.

\bibitem{Robinson-DAVIS2019}
A.~Robinson, F.~J. Lawin, M.~Danelljan, and M.~Felsberg.
\newblock Discriminative learning and target attention for the 2019 davis
  challenge on video object segmentation.
\newblock {\em CVPR Workshops}, 2019.

\bibitem{Robinson-CVPR2020}
A.~Robinson, F.~J. Lawin, M.~Danelljan, F.~S. Khan, and M.~Felsberg.
\newblock Learning fast and robust target models for video object segmentation.
\newblock In {\em CVPR}, June 2020.

\bibitem{Rother-SIGGRAPH2004}
C.~Rother, V.~Kolmogorov, and A.~Blake.
\newblock "grabcut": Interactive foreground extraction using iterated graph
  cuts.
\newblock {\em Transactions on Graphics}, 23(3), 2004.

\bibitem{Russakovsky-IJCV2015}
O.~Russakovsky, J.~Deng, H.~Su, J.~Krause, S.~Satheesh, S.~Ma, Z.~Huang,
  A.~Karpathy, A.~Khosla, M.~Bernstein, A.~C. Berg, and L.~Fei-Fei.
\newblock {ImageNet Large Scale Visual Recognition Challenge}.
\newblock {\em IJCV}, 115(3), 2015.

\bibitem{Sabour-NeurIPS2017}
S.~Sabour, N.~Frosst, and G.~E. Hinton.
\newblock Dynamic routing between capsules.
\newblock {\em NeurIPS}, 2017.

\bibitem{Sabour-2020}
S.~Sabour, A.~Tagliasacchi, S.~Yazdani, G.~E. Hinton, and D.~J. Fleet.
\newblock Unsupervised part representation by flow capsules.
\newblock {\em arXiv preprint arXiv:2011.13920}, 2020.

\bibitem{Seong-DAVIS2020}
H.~Seong, J.~Hyun, and E.~Kim.
\newblock A kernel-based approach for video object segmentation.
\newblock {\em CVPR Workshops}, 2020.

\bibitem{Shaban-DAVIS2017}
A.~Shaban, A.~Firl, A.~Humayun, J.~Yuan, X.~Wang, P.~Lei, N.~Dhanda, B.~Boots,
  J.~M. Rehg, and F.~Li.
\newblock Multiple-instance video segmentation with sequence-specific object
  proposals.
\newblock {\em CVPR Workshops}, 2017.

\bibitem{Sharir-DAVIS2017}
G.~Sharir, E.~Smolyansky, and I.~Friedman.
\newblock Video object segmentation using tracked object proposals.
\newblock {\em CVPR Workshops}, 2017.

\bibitem{Sun-DAVIS2018}
J.~Sun, D.~Yu, Y.~Li, and C.~Wang.
\newblock Mask propagation network for video object segmentation.
\newblock {\em CVPR Workshops}, 2018.

\bibitem{tmtriet-DAVIS2020}
M.-T. Tran, T.~Hoang, T.~V. Nguyen, T.-N. Le, E.~Nguyen, M.~Le, H.~Nguyen-Dinh,
  X.~Hoang, and M.~N. Do.
\newblock Multi-referenced guided instance segmentation framework for
  semi-supervised video instance segmentation.
\newblock {\em CVPR Workshops}, 2020.

\bibitem{tmtriet-DAVIS2019}
M.-T. Tran, T.-N. Le, T.~V. Nguyen, V.~Ton-That, T.-H. Hoang, N.-M. Bui, T.-L.
  Do, Q.-A. Luong, V.-T. Nguyen, D.~A. Duong, and M.~N. Do.
\newblock Guided instance segmentation framework for semi-supervised video
  instance segmentation.
\newblock In {\em CVPR Workshops}, 2019.

\bibitem{tmtriet-DAVIS2018}
M.-T. Tran, V.~Ton-That, T.-N. Le, K.-T. Nguyen, T.~V. Ninh, T.-K. Le, V.-T.
  Nguyen, T.~V. Nguyen, and M.~N. Do.
\newblock Context-based instance segmentation in video sequences.
\newblock {\em CVPR Workshops}, 2018.

\bibitem{Tsai-CVPR2016}
Y.~H. Tsai, M.~H. Yang, and M.~J. Black.
\newblock Video segmentation via object flow.
\newblock In {\em CVPR}, 2016.

\bibitem{Voigtlaender-CVPR2019}
P.~Voigtlaender, Y.~Chai, F.~Schroff, H.~Adam, B.~Leibe, and L.-C. Chen.
\newblock Feelvos: Fast end-to-end embedding learning for video object
  segmentation.
\newblock In {\em CVPR}, 2019.

\bibitem{Voigtlaender-DAVIS2017}
P.~Voigtlaender and B.~Leibe.
\newblock Online adaptation of convolutional neural networks for the 2017 davis
  challenge on video object segmentation.
\newblock {\em CVPR Workshops}, 2017.

\bibitem{Petrosyan-DAVIS2018}
V.Petrosyan, O.~{\"O}rnsberg, and A.~Proutiere.
\newblock Video object segmentation via tracking edges and classifying
  segments.
\newblock {\em CVPR Workshops}, 2018.

\bibitem{vltanh-DAVIS2020}
T.~Vu-Le, H.~Nguyen-Le, E.~Nguyen, M.~N. Do, and M.~Tran.
\newblock Video object segmentation with memory augmentation and multi-pass
  approach.
\newblock {\em CVPR Workshops}, 2020.

\bibitem{Wang-DAVIS2019}
B.~Wang, C.~Zheng, N.~Wang, S.~Wang, X.~Zhang, S.~Liu, S.~Gao, K.~Lu, D.~Zhang,
  L.~Shen, Y.~Wang, and Y.~Xu.
\newblock Object-based spatial similarity for semi-supervised video object
  segmentation.
\newblock {\em CVPR Workshops}, 2019.

\bibitem{Wang-CVPR2019}
Q.~Wang, L.~Zhang, L.~Bertinetto, W.~Hu, and P.~H. Torr.
\newblock Fast online object tracking and segmentation: A unifying approach.
\newblock In {\em CVPR}, 2019.

\bibitem{Weinzaepfel-ICCV2013}
P.~Weinzaepfel, J.~Revaud, Z.~Harchaoui, and C.~Schmid.
\newblock Deepflow: Large displacement optical flow with deep matching.
\newblock In {\em ICCV}, 2013.

\bibitem{Tong-CVPR2017}
T.~Xiao, S.~Li, B.~Wang, L.~Lin, and X.~Wang.
\newblock Joint detection and identification feature learning for person
  search.
\newblock In {\em CVPR}, 2017.

\bibitem{Xie-DAVIS2020}
H.~Xie, Y.~Huang, A.~Xu, J.~Lan, and W.~Sun.
\newblock Depth-aware space-time memory network for video object segmentation.
\newblock {\em CVPR Workshops}, 2020.

\bibitem{Xiong-CVPR2019}
Y.~Xiong, R.~Liao, H.~Zhao, R.~Hu, M.~Bai, E.~Yumer, and R.~Urtasun.
\newblock Upsnet: A unified panoptic segmentation network.
\newblock In {\em CVPR}, 2019.

\bibitem{Kai-CVPR2019}
K.~Xu, L.~Wen, G.~Li, L.~Bo, and Q.~Huang.
\newblock Spatiotemporal cnn for video object segmentation.
\newblock In {\em CVPR}, 2019.

\bibitem{Xu-BMVC2017}
N.~Xu, B.~Price, S.~Cohen, J.~Yang, and T.~Huang.
\newblock Deep grabcut for object selection.
\newblock {\em BMVC}, 2017.

\bibitem{Xu-DAVIS2018}
S.~Xu, L.~Bao, and P.~Zhou.
\newblock Class-agnostic video object segmentation without semantic
  re-identification.
\newblock {\em CVPR Workshops}, 2018.

\bibitem{Xu-CVPR2019}
S.~Xu, D.~Liu, L.~Bao, W.~Liu, and P.~Zhou.
\newblock Mhp-vos: Multiple hypotheses propagation for video object
  segmentation.
\newblock In {\em CVPR}, 2019.

\bibitem{Jinnan-IEEEAccess2021}
J.~Yan, T.-N. Le, K.-D. Nguyen, M.-T. Tran, T.-T. Do, and T.~V. Nguyen.
\newblock Mirrornet: Bio-inspired camouflaged object segmentation.
\newblock {\em IEEE Access}, 9:43290--43300, 2021.

\bibitem{Yang-CVPR2016}
J.~Yang, B.~Price, S.~Cohen, H.~Lee, and M.~H. Yang.
\newblock Object contour detection with a fully convolutional encoder-decoder
  network.
\newblock In {\em CVPR}, 2016.

\bibitem{Yang-DAVIS2020}
Z.~Yang, Y.~Ding, Y.~Wei, and Y.~Yang.
\newblock Cfbi+: Collaborative video object segmentation by multi-scale
  foreground-background integration.
\newblock {\em CVPR Workshops}, 2020.

\bibitem{Zhang-DAVIS2020}
P.~Zhang, L.~Hu, B.~Zhang, and P.~Pan.
\newblock Spatial constrained memory network for semi-supervised video object
  segmentation.
\newblock {\em CVPR Workshops}, 2020.

\bibitem{Zhang-RS2019}
W.~Zhang, P.~Tang, and L.~Zhao.
\newblock Remote sensing image scene classification using cnn-capsnet.
\newblock {\em Remote Sensing}, 11(5):494, 2019.

\bibitem{Zhao-DAVIS2017}
H.~Zhao.
\newblock Some promising ideas about multi-instance video segmentation.
\newblock {\em CVPR Workshops}, 2017.

\bibitem{Zhao-CVPR2017}
H.~Zhao, J.~Shi, X.~Qi, X.~Wang, and J.~Jia.
\newblock Pyramid scene parsing network.
\newblock In {\em CVPR}, 2017.

\bibitem{Zhou-TPAMI2017}
B.~Zhou, A.~Lapedriza, A.~Khosla, A.~Oliva, and A.~Torralba.
\newblock Places: A 10 million image database for scene recognition.
\newblock {\em Transactions on Pattern Analysis and Machine Intelligence},
  2017.

\end{thebibliography}
}

\end{document}